\documentclass{article}

\PassOptionsToPackage{numbers,sort&compress}{natbib}
\usepackage[final]{fragflex_ai4dd}
\usepackage{graphicx}

\usepackage{booktabs}            % professional-quality tables
\usepackage{multirow}            % tabular cells spanning multiple rows
\usepackage{amsfonts}            % blackboard math symbols
\usepackage{amsmath,amssymb}
\usepackage{adjustbox}
\usepackage{caption}
\usepackage{algorithm}
\usepackage{float}
\usepackage[noend]{algpseudocode}

\usepackage[utf8]{inputenc} % allow utf-8 input
\usepackage[T1]{fontenc}    % use 8-bit T1 fonts
\usepackage{hyperref}       % hyperlinks
\usepackage{url}            % simple URL typesetting
\usepackage{booktabs}       % professional-quality tables
\usepackage{amsfonts}       % blackboard math symbols
\usepackage{nicefrac}       % compact symbols for 1/2, etc.
\usepackage{microtype}      % microtypography
\usepackage{xcolor}         % colors

\title{FragFlex: Fragment-Based Flexible Molecular Generation
for Exploring Vast Chemical Space}
\workshoptitle{NeurIPS 2026 Workshop on AI for Drug Discovery}
\makeatletter
\renewcommand{\@noticestring}{%
  Accepted at the NeurIPS 2026 Workshop on AI for Drug Discovery:
  Bridging the Translation Gap.%
}
\makeatother

\author{%
  Sihyun Park \\
  Carnegie Mellon University \\
  \texttt{sihyunpa@andrew.cmu.edu}
  \And
  David Ryan Koes \\
  University of Pittsburgh \\
  \texttt{dkoes@pitt.edu}
}

\begin{document}

\maketitle

\begin{abstract}
Drug discovery requires efficient exploration of an astronomically large chemical space. Molecular graph diffusion models offer a promising approach to this task, but many require graph size to be fixed before generation. Although atom-level insertion and deletion can enable flexible-size generation, jointly modeling molecular size, atom types, and bonds while maintaining chemical validity remains challenging. We introduce FragFlex, a flexible-length discrete diffusion model that performs node insertion and deletion at the fragment level. During denoising, FragFlex jointly learns the size, fragment identities, and inter-fragment connectivity of the fragment graph. A neural attachment-site predictor and valence-aware assembler then convert the generated fragment graph into an atom-level molecule. In large-scale experiments generating 10 million molecules, FragFlex achieves near-perfect chemical validity while maintaining high uniqueness, demonstrating a strong balance between the two.
\end{abstract}

\section{Introduction}
Drug discovery requires exploring an astronomically large chemical space. Estimates of the size of drug-like chemical space vary substantially with the underlying molecular constraints and enumeration strategies, ranging from approximately $10^{33}$ molecules \citep{polishchuk2013estimation} to $10^{60}$ molecules \citep{bohacek1996art}. Efficiently searching this space is therefore key to reducing the computational and experimental costs of finding promising drug candidates \citep{schneider2020rethinking, thomas2026generative}.

De novo molecular generation offers a promising strategy for exploring chemical space by directly proposing novel molecular structures. Deep generative models can learn molecular distributions from existing compound datasets and generate new molecules. Notably, denoising diffusion models have driven substantial progress across
image generation \citep{ho2020denoising}, discrete text generation
\citep{austin2021structured,lou2023discrete}, and molecular graph generation
\citep{vignac2022digress}.

However, many existing molecular graph diffusion models keep the number of nodes fixed throughout denoising, requiring molecular size to be determined before generation begins. Attempts to overcome fixed-length generation are not unique to molecular graphs: masked diffusion models (MDMs) \citep{sahoo2024simple} have been extended to support flexible-length text generation through mask-token insertion \citep{kim2026any}. For molecular graphs, Ninniri et al. proposed GrIDDD \citep{ninniri2026graph}, allowing atom nodes to be inserted and deleted during denoising. Despite this flexibility, maintaining high chemical validity remains challenging when molecular graph size is allowed to vary through atom-level insertion and deletion. This difficulty arises because the model must jointly learn molecular size, atom types, and bond assignments while satisfying strict chemical constraints, such as atomic valence.

Fragment-based molecular generation offers a natural way to mitigate this challenge. Instead of constructing every ring and functional group atom by atom, these methods assemble reusable substructures, reducing invalid intermediates \citep{jin2018junction, jin2020hierarchical}. Fragment-level diffusion and flow models further improve efficiency and scalability by denoising a coarser graph \citep{lee2026fragfm}. In particular, Levy
and Rector-Brooks proposed FragDiffusion \citep{levy2023molecular}, extending discrete graph diffusion from atoms to fragments.  Nevertheless, the two capabilities have largely developed separately: flexible-size diffusion typically operates on atoms or generic tokens, whereas representative fragment diffusion models initialize a fragment graph whose size is sampled before denoising.

To bridge this gap, we introduce FragFlex, a flexible-length discrete diffusion model that performs node insertion and deletion at the fragment level. The model jointly learns fragment count, fragment identities, and inter-fragment connectivity while dynamically adjusting the graph size during denoising. This design allows molecular size and composition to co-evolve within a single generative trajectory while retaining the chemical inductive bias of fragment-based generation. Finally, a neural attachment-site predictor and valence-aware assembler convert the generated fragment graph into an atom-level molecule. In large-scale experiments generating 10 million molecules, FragFlex maintains both high uniqueness and chemical validity, showing a favorable trade-off between the two.

Our main contributions are as follows:
\begin{itemize}
\item We introduce FragFlex, a discrete diffusion model that enables flexible molecular-size generation through fragment-level node insertion and deletion.

\item We develop a neural attachment-site predictor and valence-aware assembler for converting fragment graphs into valid molecules.

\item In experiments generating 10 million molecules, FragFlex achieves high chemical validity and a favorable uniqueness--validity trade-off.

\end{itemize}

\section{Related Work}
\label{app:related_work}

\subsection{De Novo Molecular Generation}

De novo molecular generation aims to learn distributions over molecular
structures for unconditional or property-guided sampling \citep{pang2024deep}. Early neural
approaches often represented molecules as SMILES strings and applied recurrent
models or variational autoencoders \citep{gomez2018automatic, segler2018generating, kusner2017grammar}. Graph-based
methods instead operate directly on atoms and bonds, using variational
autoencoders, autoregressive models, and normalizing flows
\citep{simonovsky2018graphvae, liu2018constrained,shi2020graphaf,luo2021graphdf}.

More recently, score-based and diffusion models have formulated molecular
generation as iterative denoising of molecular graphs
\citep{jo2022score,jo2023graph}. In particular, discrete diffusion methods
such as D3PM \citep{austin2021structured} enable diffusion over categorical
variables, providing a natural framework for molecular graphs with discrete
atom and bond types. DiGress applies this framework to graph generation using
categorical transition kernels and a permutation-equivariant graph transformer
\citep{vignac2022digress}.

\subsection{Flexible-Length Molecular Generation}

Autoregressive models can naturally generate molecules of varying sizes through
a stopping decision, whereas many graph diffusion models sample the graph size
beforehand and keep it fixed during denoising. DiGress, for instance, samples
the graph size from the training-set size distribution before denoising and
keeps it fixed throughout generation \citep{vignac2022digress}.

Recent methods instead allow graph size to change during generation. IFH uses
insertion, filling, and halting operations \citep{cognolato2024ifh}, while
GrIDDD deletes nodes during forward diffusion and inserts them during reverse
generation \citep{ninniri2026graph}. In the sequence domain, FlexMDM enables
flexible-length diffusion through mask-token insertion \citep{kim2026any}.

FragFlex applies this size-adaptive principle at the fragment level. Starting
from a single root fragment, it predicts fragment insertions and jointly
denoises fragment identities and connectivity. 

\subsection{Fragment-Based Molecular Generation}

Fragment-based molecular generative models construct molecules from reusable
molecular substructures, or fragments, rather than individual atoms. This
coarse representation can reduce the number of generative decisions while
incorporating chemically meaningful substructures into the generation process
\citep{voloboev2024review,jinsong2024molecular}. JT-VAE generates a junction
tree over molecular substructures before assembling the molecular graph
\citep{jin2018junction}, while HierVAE extends this approach with hierarchical
motifs and coarse-to-fine generation \citep{jin2020hierarchical}.

Fragment-level denoising methods instead operate directly on coarse fragment
graphs. FragDiffusion applies discrete diffusion to BRICS-derived fragment graphs
whose size is selected before denoising and reconstructs atom-level molecules
using attachment modes stored in a precomputed lookup table
\citep{levy2023molecular}. FragFM
combines fragment-level discrete flow matching with coarse-to-fine atom-level
reconstruction \citep{lee2026fragfm}.

FragFlex combines fragment-level generation with size-adaptive diffusion.
Unlike prior fragment-level denoising methods, it allows the number of fragments
to change during generation while jointly modeling fragment identities and
connectivity. It further predicts attachment sites directly and uses a
valence-aware assembler to reconstruct the final atom-level molecule.

\begin{figure}
  \centering
\includegraphics[
    width=1.0\linewidth,
    trim=100 170 90 150,
    clip
]{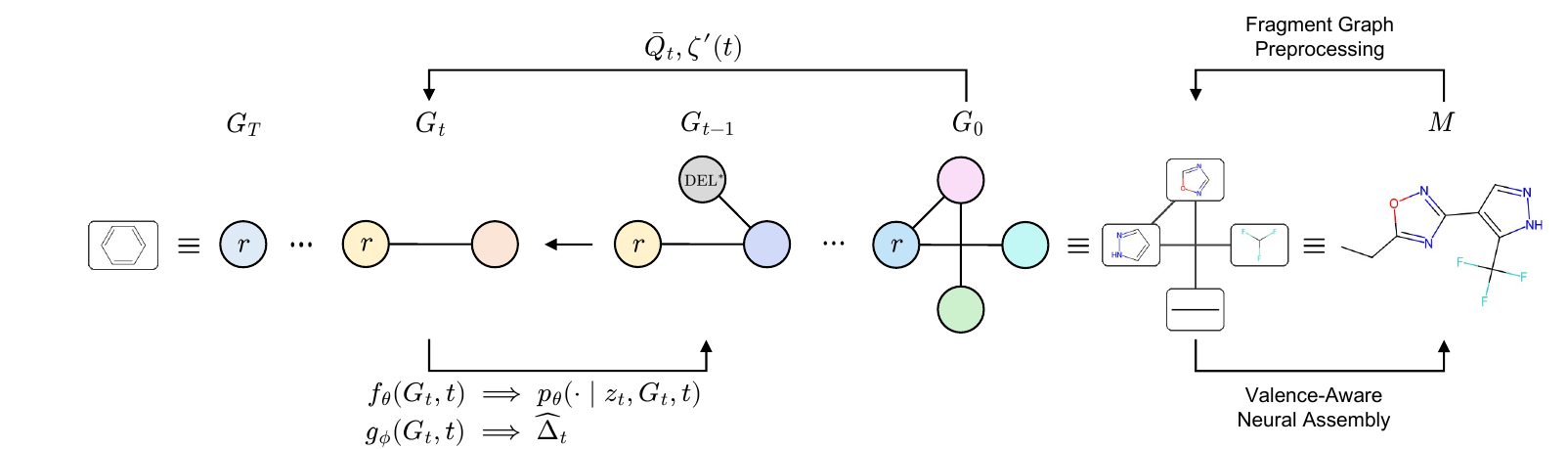}
\caption{
Overview of FragFlex.
An atom-level molecule $M$ is preprocessed into a clean fragment graph
$G_0$. Forward diffusion applies categorical noise $\bar{Q}_t$ and
structural deletion $\zeta'(t)$ while protecting one root, yielding
the single-fragment state $G_T$. During sampling, $g_\phi$ predicts the
insertion count $\widehat{\Delta}_t$, and $f_\theta$ predicts the node
types and edge connections of the added $\textsc{del}^{*}$
placeholders.
At the final denoising step, predicted attachment sites
guide the valence-aware assembly of the final molecule.
}
\label{fig:fragflex_overview}
  \vspace{-5pt}
\end{figure}

\section{Methods}

\textbf{Overview.} \quad
FragFlex generates a molecule in two stages (Figure~\ref{fig:fragflex_overview}).
First, starting from a single root fragment, a discrete diffusion model generates a coarse graph by jointly modeling fragment identities, connectivity, and graph size (Figure~\ref{fig:molgen}). Second, a neural attachment-site predictor and a valence-aware
assembler convert this fragment graph into an atom-level molecular graph.
Our model combines FragDiffusion~\citep{levy2023molecular}, a discrete diffusion model over fragment graphs, with GrIDDD~\citep{ninniri2026graph}, an atom-level diffusion model that enables variable graph sizes through node insertions and deletions.

\textbf{Problem definition.} \quad
Let $\mathcal V=\{F_1,\ldots,F_K\}$ be a fixed vocabulary of BRICS-derived
fragment types extracted from the training molecules \citep{degen2008art}. Each $F_k$ specifies an
atom-and-bond substructure and may contain either a single heavy atom or multiple
heavy atoms. We retain the $K=100$ most frequent fragment types. We represent an atom-level molecule $M$ by a clean fragment graph $G_0=(X_0,E_0,S_0)$ with $N$ fragments. The node label $X_{0,i}\in\{1,\ldots,K\}$ identifies the fragment at node $i$, and $E_{0,ij}\in\{0,1\}$ indicates whether fragments $i$ and $j$ are connected. For each connected fragment pair $(i,j)$, $S_{0,ij}$ and $S_{0,ji}$ denote the local indices of the attachment atoms in fragments $i$ and $j$, respectively, with $S_{0,ij}\in\{0,\ldots,A_i-1\}$, where $A_i$ is the number of atoms in fragment $i$. Our goal is to jointly model fragment identities, topology, attachment sites, and graph size. Unlike FragDiffusion~\citep{levy2023molecular}, which assembles molecules using a fragment-pair lookup table precomputed from attachment frequencies in the training data, FragFlex represents fragment connectivity with binary edges and directly predicts the corresponding attachment atoms.

\textbf{Training.} \quad
Given a clean fragment graph $G_0=(X_0,E_0,S_0)$ with node set $V_0=\{1,\ldots,N\}$, we sample $t\sim\mathrm{Unif}\{1,\ldots,T\}$ and a root $r\sim\mathrm{Unif}(V_0)$.
Each non-root fragment is assigned a deletion time $\tau_i\sim\mathrm{Cat}(\zeta')$, while we set $\tau_r=T+1$ so that the root is never deleted. Following GrIDDD~\citep{ninniri2026graph}, $\zeta'(t)$ is a normalized logistic event density.
At time $t$, fragments with $\tau_i<t$ are removed and boundary fragments with $\tau_i=t$ are represented by the transient state $\textsc{del}^{*}$; their incident edges are marked identically.
The retained fragment and edge labels are corrupted using the
empirical-marginal kernel
\begin{equation}
\bar{Q}_t(m)
=
\bar{\alpha}_t I
+
(1-\bar{\alpha}_t)\mathbf{1}m^\top,
\label{eq:ordinary_kernel}
\end{equation}
where $m$ is the corresponding training marginal and $\bar{\alpha}_t$
follows a cosine schedule.
At $t=T$, the terminal state therefore contains one protected fragment drawn from the empirical fragment marginal.

An edge-aware graph transformer \citep{yun2019graph} $f_\theta$ serves as the main denoiser and predicts clean fragment logits $Z^X$, binary edge logits $Z^E$, and directed attachment-site logits $Z^S$ from $G_t$. Its encoder combines time and root embeddings with the cycle features of DiGress~\citep{vignac2022digress}. A separate count model $g_\phi$ receives $G_t$ after removing the $\textsc{del}^{\ast}$ fragments and predicts their number $\Delta_t$.
We optimize the joint objective
\begin{equation}
\mathcal{L}
=
\mathcal{L}_X
+
\lambda_E\mathcal{L}_E
+
\lambda_\Delta\mathcal{L}_\Delta
+
\lambda_S\mathcal{L}_S,
\label{eq:training_loss}
\end{equation}
where all terms are cross-entropies. These four terms supervise fragment identities, fragment edges,
deletion counts, and attachment sites, respectively.

\textbf{Fragment graph sampling.} \quad
Starting from a single root fragment at \(t=T\), the reverse process
constructs a fragment graph until \(t=1\).
At each reverse step, $g_\phi$ predicts how many $\textsc{del}^{*}$ placeholders to append.
The denoiser $f_\theta$ outputs clean node and edge logits, which
parameterize the reverse distribution
\begin{equation}
 p_\theta(z_{t-1}\mid z_t,G_t,t)
 =\sum_{z_0}q(z_{t-1}\mid z_t,z_0)
 p_\theta(z_0\mid G_t,t).
\end{equation}
Since attachment sites are needed only for final molecular assembly, we evaluate the site head only at \(t=1\), avoiding unnecessary computation during reverse diffusion. The resulting probabilities \(\widehat{P}^{S}\) are passed to the atom-level assembler.

\textbf{Atom-level assembly.} \quad
After masking infeasible attachment atoms and renormalizing
$\widehat{P}^{S}$ to $\widetilde{P}^{S}$, the assembler considers the
top-$k$ atoms for each generated fragment edge $(i,j)$ and scores
candidate pairs $(a,b)$ by
\begin{equation}
\ell_{ij}(a,b)
=
\log \widetilde{P}^{S}_{ij}(a)
+
\log \widetilde{P}^{S}_{ji}(b).
\label{eq:site_pair_score}
\end{equation}

Beam search processes the fragment edges sequentially, adding one
candidate single bond at each step and discarding duplicate bonds or
candidates that violate valence constraints under RDKit \citep{landrum2023rdkit}.
The assembler keeps up to a fixed number of distinct valid molecules
and selects one uniformly at random. Thus, chemical
constraints filter the learned site proposals without requiring an
attachment lookup table at inference.

\section{Results}

\begin{table}[t]
\centering
\caption{
Validity and uniqueness across generation scales.
$|V\cap U|$ denotes the number of valid and unique molecules obtained
from $10^7$ generated samples. QED and SA are evaluated at $n=10^5$.
}
\label{tab:large_scale_generation}

\begingroup
\footnotesize
\setlength{\tabcolsep}{2.5pt}
\renewcommand{\arraystretch}{1.08}

\begin{adjustbox}{max width=\linewidth,center}
\begin{tabular}{@{}lcc*{9}{r}@{}}
\toprule
& & &
\multicolumn{3}{c}{Validity (\%) $\uparrow$} &
\multicolumn{3}{c}{Uniqueness (\%) $\uparrow$} &
\multicolumn{1}{c}{QED $\uparrow$} &
\multicolumn{1}{c}{SA $\downarrow$} &
\multicolumn{1}{c}{$|V\cap U|$ $\uparrow$} \\

\cmidrule(lr){4-6}
\cmidrule(lr){7-9}
\cmidrule(lr){10-11}
\cmidrule(l){12-12}

Model &
Unit &
Ins./del. &
$10^5$ & $10^6$ & $10^7$ &
$10^5$ & $10^6$ & $10^7$ &
\multicolumn{2}{c}{$n=10^5$} &
$n=10^7$ \\

\midrule

DiGress
    & Atom & No
    & 71.47 & 71.33 & 71.36
    & \textbf{99.99} & \textbf{99.95} & \textbf{99.68}
    & 0.733 & 3.702
    & 7,113,016 \\

GrIDDD
    & Atom & Yes
    & 88.10 & 88.23 & 88.22
    & 99.83 & 99.04 & 97.10
    & 0.651 & \underline{3.413}
    & 8,566,049 \\

IFH
    & Atom & Yes
    & 57.98 & 57.87 & 57.92
    & \underline{99.98} & \underline{99.85}
    & \underline{99.40}
    & 0.725 & 5.367
    & 5,757,629 \\

\midrule

JT-VAE
    & Fragment & No
    & \textbf{99.73} & \textbf{99.74} & \textbf{99.75}
    & 99.25 & 96.72 & 89.40
    & \underline{0.762} & 3.418
    & 8,916,975 \\

FragDiffusion
    & Fragment & No
    & 96.25 & 96.16 & 96.18
    & 99.81 & 99.07 & 96.33
    & 0.609 & 6.840
    & \underline{9,265,827} \\

\textbf{FragFlex}
    & Fragment & Yes
    & \underline{99.67} & \underline{99.67}
    & \underline{99.67}
    & 99.94 & 99.37 & 95.52
    & \textbf{0.764} & \textbf{3.021}
    & \textbf{9,520,600} \\

\bottomrule
\end{tabular}
\end{adjustbox}

\endgroup
\end{table}

\textbf{Dataset and experiments.} \quad
The evaluated models use datasets derived from ZINC \citep{sterling2015zinc}.
For each baseline, we retain the default dataset and model
configuration specified in its original implementation.
FragFlex follows the preprocessing pipeline of
FragDiffusion~\cite{levy2023molecular}, using BRICS fragmentation
and fragment graph construction.

Conventional molecular generation evaluations typically consider
tens to hundreds of thousands of generated molecules.
To assess chemical-space exploration at a substantially larger scale,
we generate up to $10^7$ molecules from each model.
All models are evaluated at the same sample sizes using identical
metric definitions.
Validity is the fraction of generated molecules accepted by RDKit,
while uniqueness is computed over canonicalized valid molecules.
\begin{figure}[t]
\centering
\captionsetup{skip=10pt}

\begin{minipage}[t]{0.45\linewidth}
    \centering
    \includegraphics[
        width=\linewidth,
        height=0.145\textheight,
        keepaspectratio,
        trim=270 120 270 120,
        clip
    ]{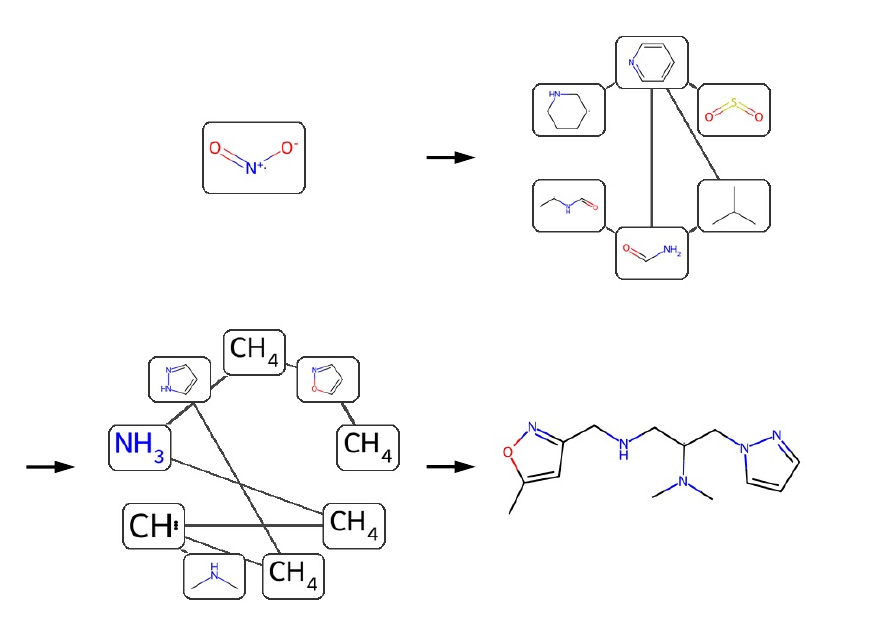}
    \captionof{figure}{FragFlex sampling steps.}
    \label{fig:molgen}
\end{minipage}
\hfill
\begin{minipage}[t]{0.54\linewidth}
    \centering
    \includegraphics[
        width=\linewidth,
        height=0.145\textheight,
        keepaspectratio,
        trim=0 15 0 15,
        clip
    ]{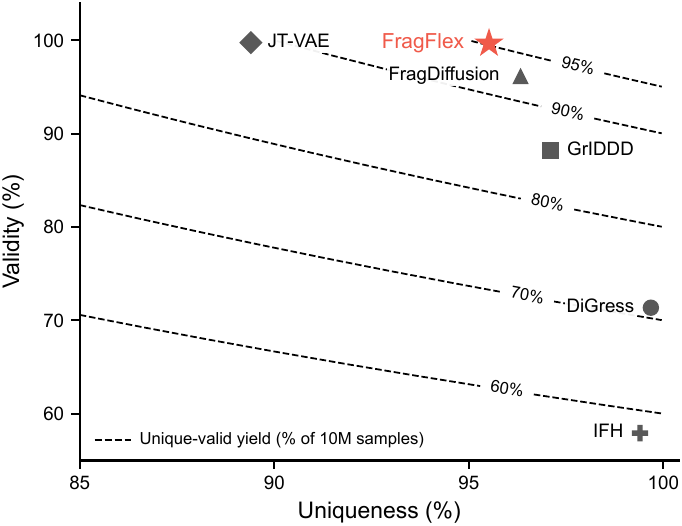}
    \captionof{figure}{Validity--uniqueness trade-off at $10^7$ samples.}
    \label{fig:validity_uniqueness}
\end{minipage}

\vspace{-12pt}
\end{figure}

\textbf{Large-scale molecular generation.} \quad
Table~\ref{tab:large_scale_generation} reports validity and uniqueness as the number
of generated molecules increases from $10^5$ to $10^7$.
We omit novelty from the main comparison because it is nearly saturated
at 100\% across all evaluated models and therefore provides little
discriminative value.
Across all models, validity remains largely unaffected by the number of
generated samples. In contrast, uniqueness gradually decreases as
repeated molecules accumulate at larger sampling scales.

Fragment-based models consistently achieve substantially higher validity than
atom-based models, although they generally exhibit lower uniqueness at $10^7$
samples. JT-VAE achieves the highest validity at every evaluated scale but also
exhibits the largest decline in uniqueness as the sampling scale increases.
FragDiffusion shows the lowest validity among the fragment-based models.
FragFlex approaches JT-VAE's high validity while retaining uniqueness
comparable to FragDiffusion, thereby combining the strengths of the two
fragment-based baselines. In addition, at $n=10^5$, FragFlex achieves the
highest QED \citep{bickerton2012quantifying} score ($0.764$) and the lowest SA \citep{ertl2009estimation} score ($3.021$) across all
evaluated atom- and fragment-based models.

Figure~\ref{fig:validity_uniqueness} highlights the relationship
between validity and uniqueness at $10^7$ samples.
FragFlex maintains $99.67\%$ validity across all three scales while
retaining $95.52\%$ uniqueness at $10^7$ samples.
This balance results in $9{,}520{,}600$ distinct valid molecules, the
highest $|V\cap U|$ count among all models.
FragFlex lies closest to the upper-right corner and is the only model above the $95\%$ unique-valid yield contour, demonstrating the
strongest balance between validity and uniqueness at scale.

\section{Conclusion}
We introduced FragFlex, a flexible-length molecular diffusion model
that performs node insertion and deletion at the fragment level.
FragFlex jointly predicts the number, identities, and connectivity of
molecular fragments.
A neural attachment-site predictor and valence-aware assembler then
convert the generated fragment graph into an atom-level molecule.
Together, these components allow molecular size to evolve throughout
generation while retaining structural information encoded by
fragments.

At a scale of $10^7$ generated molecules, FragFlex produces more than
$9.5$ million distinct valid molecules, demonstrating a strong balance
between validity and uniqueness.
These results suggest that fragment-level structural flexibility
provides an effective foundation for large-scale chemical-space
exploration.
This flexibility also suggests a promising extension to reinforcement
learning, where property-based rewards could balance the exploration
of new fragment combinations with the exploitation of promising
molecules.

The effectiveness of size-adaptive diffusion models such as GrIDDD
in molecular property targeting motivates extending FragFlex to
protein-target-conditioned generation.
This extension would allow fragment composition and molecular size
to adapt to target-specific binding environments.
Future work will also investigate property-guided optimization and
downstream drug discovery applications.

\clearpage
\bibliographystyle{plainnat}
\bibliography{fragflex_ai4dd}

@article{polishchuk2013estimation,
  title={Estimation of the size of drug-like chemical space based on {GDB}-17 data},
  author={Polishchuk, Pavel G and Madzhidov, Timur I and Varnek, Alexandre},
  journal={Journal of Computer-Aided Molecular Design},
  volume={27},
  number={8},
  pages={675--679},
  year={2013},
  publisher={Springer}
}

@article{bohacek1996art,
  title={The art and practice of structure-based drug design: a molecular modeling perspective},
  author={Bohacek, Regine S and McMartin, Colin and Guida, Wayne C},
  journal={Medicinal Research Reviews},
  volume={16},
  number={1},
  pages={3--50},
  year={1996},
  publisher={Wiley}
}

@inproceedings{kim2026any,
  title={Any-order flexible length masked diffusion},
  author={Kim, Jaeyeon and Kit, Lee and Domingo i Enrich, Carles and Du, Yilun and Kakade, Sham and Ngotiaoco, Timothy and Chen, Sitan and Albergo, Michael},
  booktitle={International Conference on Learning Representations},
  year={2026}
}

@inproceedings{ninniri2026graph,
  title={Graph diffusion that can insert and delete},
  author={Ninniri, Matteo and Podda, Marco and Bacciu, Davide},
  booktitle={Advances in Neural Information Processing Systems},
  volume={38},
  pages={78375--78401},
  year={2025},
  doi={10.52202/085713-2625}
}

@inproceedings{levy2023molecular,
  title={Molecular fragment-based diffusion model for drug discovery},
  author={Levy, Daniel and Rector-Brooks, Jarrid},
  booktitle={ICLR 2023 Workshop on Machine Learning for Drug Discovery},
  year={2023}
}

@article{vignac2022digress,
  title={{DiGress}: Discrete denoising diffusion for graph generation},
  author={Vignac, Cl{\'e}ment and Krawczuk, Igor and Siraudin, Antoine and Wang, Bohan and Cevher, Volkan and Frossard, Pascal},
  journal={arXiv preprint arXiv:2209.14734},
  year={2022}
}

@inproceedings{cognolato2024ifh,
  title={{IFH}: A diffusion framework for flexible design of graph generative models},
  author={Cognolato, Samuel and Sperduti, Alessandro and Serafini, Luciano},
  booktitle={ECAI 2024: 27th European Conference on Artificial Intelligence},
  pages={3039--3046},
  year={2024}
}

@inproceedings{jin2018junction,
  title={Junction tree variational autoencoder for molecular graph generation},
  author={Jin, Wengong and Barzilay, Regina and Jaakkola, Tommi},
  booktitle={International Conference on Machine Learning},
  pages={2323--2332},
  year={2018},
  organization={PMLR}
}

@inproceedings{jin2020hierarchical,
  title={Hierarchical generation of molecular graphs using structural motifs},
  author={Jin, Wengong and Barzilay, Regina and Jaakkola, Tommi},
  booktitle={International Conference on Machine Learning},
  pages={4839--4848},
  year={2020},
  organization={PMLR}
}

@article{sterling2015zinc,
  title={{ZINC} 15--Ligand discovery for everyone},
  author={Sterling, Teague and Irwin, John J},
  journal={Journal of Chemical Information and Modeling},
  volume={55},
  number={11},
  pages={2324--2337},
  year={2015},
  publisher={ACS Publications}
}

@article{landrum2023rdkit,
  title={{RDKit}/{RDKit}: 2023\_03\_1 ({Q1} 2023) release},
  author={Landrum, Greg and Tosco, Paolo and Kelley, Brian and Cosgrove, David and Vianello, Riccardo and Kawashima, Eisuke and Jones, Gareth and Dalke, Andrew and Cole, Brian and Swain, Matt and others},
  journal={Zenodo},
  year={2023}
}

@article{gomez2018automatic,
  title={Automatic chemical design using a data-driven continuous representation of molecules},
  author={G{\'o}mez-Bombarelli, Rafael and Wei, Jennifer N and Duvenaud, David and Hern{\'a}ndez-Lobato, Jos{\'e} Miguel and S{\'a}nchez-Lengeling, Benjam{\'i}n and Sheberla, Dennis and Aguilera-Iparraguirre, Jorge and Hirzel, Timothy D and Adams, Ryan P and Aspuru-Guzik, Al{\'a}n},
  journal={ACS Central Science},
  volume={4},
  number={2},
  pages={268},
  year={2018}
}

@inproceedings{simonovsky2018graphvae,
  title={{GraphVAE}: Towards generation of small graphs using variational autoencoders},
  author={Simonovsky, Martin and Komodakis, Nikos},
  booktitle={International Conference on Artificial Neural Networks},
  pages={412--422},
  year={2018},
  organization={Springer}
}

@article{shi2020graphaf,
  title={{GraphAF}: A flow-based autoregressive model for molecular graph generation},
  author={Shi, Chence and Xu, Minkai and Zhu, Zhaocheng and Zhang, Weinan and Zhang, Ming and Tang, Jian},
  journal={arXiv preprint arXiv:2001.09382},
  year={2020}
}

@inproceedings{luo2021graphdf,
  title={{GraphDF}: A discrete flow model for molecular graph generation},
  author={Luo, Youzhi and Yan, Keqiang and Ji, Shuiwang},
  booktitle={International Conference on Machine Learning},
  pages={7192--7203},
  year={2021},
  organization={PMLR}
}

@article{liu2018constrained,
  title={Constrained graph variational autoencoders for molecule design},
  author={Liu, Qi and Allamanis, Miltiadis and Brockschmidt, Marc and Gaunt, Alexander},
  journal={Advances in Neural Information Processing Systems},
  volume={31},
  year={2018}
}

@article{pang2024deep,
  title={Deep generative models in de novo drug molecule generation},
  author={Pang, Chao and Qiao, Jianbo and Zeng, Xiangxiang and Zou, Quan and Wei, Leyi},
  journal={Journal of Chemical Information and Modeling},
  volume={64},
  number={7},
  pages={2174--2194},
  year={2024},
  publisher={ACS Publications}
}

@inproceedings{jo2022score,
  title={Score-based generative modeling of graphs via the system of stochastic differential equations},
  author={Jo, Jaehyeong and Lee, Seul and Hwang, Sung Ju},
  booktitle={International Conference on Machine Learning},
  pages={10362--10383},
  year={2022},
  organization={PMLR}
}

@article{yun2019graph,
  title={Graph transformer networks},
  author={Yun, Seongjun and Jeong, Minbyul and Kim, Raehyun and Kang, Jaewoo and Kim, Hyunwoo J},
  journal={Advances in Neural Information Processing Systems},
  volume={32},
  year={2019}
}

@article{segler2018generating,
  title={Generating focused molecule libraries for drug discovery with recurrent neural networks},
  author={Segler, Marwin HS and Kogej, Thierry and Tyrchan, Christian and Waller, Mark P},
  journal={ACS Central Science},
  volume={4},
  number={1},
  pages={120--131},
  year={2018},
  publisher={ACS Publications}
}

@article{jo2023graph,
  title={Graph generation with diffusion mixture},
  author={Jo, Jaehyeong and Kim, Dongki and Hwang, Sung Ju},
  journal={arXiv preprint arXiv:2302.03596},
  year={2023}
}

@article{austin2021structured,
  title={Structured denoising diffusion models in discrete state-spaces},
  author={Austin, Jacob and Johnson, Daniel D and Ho, Jonathan and Tarlow, Daniel and van den Berg, Rianne},
  journal={Advances in Neural Information Processing Systems},
  volume={34},
  pages={17981--17993},
  year={2021}
}

@inproceedings{kusner2017grammar,
  title={Grammar variational autoencoder},
  author={Kusner, Matt J and Paige, Brooks and Hern{\'a}ndez-Lobato, Jos{\'e} Miguel},
  booktitle={International Conference on Machine Learning},
  pages={1945--1954},
  year={2017},
  organization={PMLR}
}

@article{voloboev2024review,
  title={A review on fragment-based de novo {2D} molecule generation},
  author={Voloboev, Sergei},
  journal={arXiv preprint arXiv:2405.05293},
  year={2024}
}

@article{jinsong2024molecular,
  title={Molecular fragmentation as a crucial step in the {AI}-based drug development pathway},
  author={Jinsong, Shao and Qifeng, Jia and Xing, Chen and Hao, Yajie and Wang, Li},
  journal={Communications Chemistry},
  volume={7},
  number={1},
  pages={20},
  year={2024},
  publisher={Nature Publishing Group}
}

@inproceedings{lee2026fragfm,
  title={{FragFM}: Hierarchical framework for efficient molecule generation via fragment-level discrete flow matching},
  author={Lee, Joongwon and Kim, Seonghwan and Moon, Seokhyun and Kim, Hyunwoo and Kim, Woo Youn},
  booktitle={International Conference on Learning Representations},
  year={2026}
}

@article{bemis1996properties,
  title={The properties of known drugs. 1. Molecular frameworks},
  author={Bemis, Guy W and Murcko, Mark A},
  journal={Journal of Medicinal Chemistry},
  volume={39},
  number={15},
  pages={2887--2893},
  year={1996},
  publisher={ACS Publications}
}

@article{schneider2020rethinking,
  title={Rethinking drug design in the artificial intelligence era},
  author={Schneider, Petra and Walters, W Patrick and Plowright, Alleyn T and Sieroka, Norman and Listgarten, Jennifer and Goodnow Jr, Robert A and Fisher, Jasmin and Jansen, Johanna M and Duca, Jos{\'e} S and Rush, Thomas S and others},
  journal={Nature Reviews Drug Discovery},
  volume={19},
  number={5},
  pages={353--364},
  year={2020},
  publisher={Nature Publishing Group}
}

@article{lou2023discrete,
  title={Discrete diffusion modeling by estimating the ratios of the data distribution},
  author={Lou, Aaron and Meng, Chenlin and Ermon, Stefano},
  journal={arXiv preprint arXiv:2310.16834},
  year={2023}
}

@article{ho2020denoising,
  title={Denoising diffusion probabilistic models},
  author={Ho, Jonathan and Jain, Ajay and Abbeel, Pieter},
  journal={Advances in Neural Information Processing Systems},
  volume={33},
  pages={6840--6851},
  year={2020}
}

@article{degen2008art,
  title={On the art of compiling and using `drug-like' chemical fragment spaces},
  author={Degen, J{\"o}rg and Wegscheid-Gerlach, Christof and Zaliani, Andrea and Rarey, Matthias},
  journal={ChemMedChem},
  volume={3},
  number={10},
  pages={1503--1507},
  year={2008},
  doi={10.1002/cmdc.200800178}
}

@article{thomas2026generative,
  title={Generative virtual screening: from search to generate and back},
  author={Thomas, Morgan and Astudillo, Raul and Raouf, Yasir S and Navarro, Carles and De Fabritiis, Gianni and Bender, Andreas},
  journal={Digital Discovery},
  year={2026},
  publisher={The Royal Society of Chemistry}
}

@article{bickerton2012quantifying,
  title={Quantifying the chemical beauty of drugs},
  author={Bickerton, G Richard and Paolini, Gaia V and Besnard, J{\'e}r{\'e}my and Muresan, Sorel and Hopkins, Andrew L},
  journal={Nature Chemistry},
  volume={4},
  number={2},
  pages={90--98},
  year={2012},
  publisher={Nature Publishing Group}
}

@article{ertl2009estimation,
  title={Estimation of synthetic accessibility score of drug-like molecules based on molecular complexity and fragment contributions},
  author={Ertl, Peter and Schuffenhauer, Ansgar},
  journal={Journal of Cheminformatics},
  volume={1},
  number={1},
  pages={8},
  year={2009},
  publisher={Springer}
}

@article{sahoo2024simple,
  title={Simple and effective masked diffusion language models},
  author={Sahoo, Subham S and Arriola, Marianne and Schiff, Yair and Gokaslan, Aaron and Marroquin, Edgar and Chiu, Justin T and Rush, Alexander and Kuleshov, Volodymyr},
  journal={Advances in Neural Information Processing Systems},
  volume={37},
  pages={130136--130184},
  year={2024}
}

%%%%%%%%%%%%%%%%%%%%%%%%%%%%%%%%%%%%%%%%%%%%%%%%%%%%%%%%%%%%

\appendix

\section{Detailed Methods}
\label{app:detailed_methods}

\subsection{Notation}
\label{app:notation}

\begin{table}[H]
    \centering
    \caption{Notation used in the main text and Appendix~\ref{app:detailed_methods}.}
    \label{tab:appendix_notation}
    \small
    \begin{tabular}{@{}p{0.19\linewidth}p{0.75\linewidth}@{}}
        \toprule
        Symbol & Meaning \\
        \midrule
        $M$ & Atom-level molecule. \\
        $\mathcal{V}=\{F_1,\ldots,F_K\}$ & Fixed vocabulary of $K$ molecular fragments. \\
        $G_0=(X_0,E_0,S_0)$ & Clean fragment graph with fragment labels, binary fragment edges, and directed attachment sites. \\
        $G_t$ & Fragment graph after structural deletion and categorical corruption at timestep $t$. \\
        $V_0=\{1,\ldots,N\}$ & Node set of the clean fragment graph. \\
        $A_i$ & Number of atoms in the fragment assigned to node $i$. \\
        $r$ & Uniformly selected root node that is protected from structural deletion. \\
        $\tau_i$ & Structural deletion time assigned to non-root node $i$. \\
        $\zeta'(t)$ & Normalized probability mass of a structural event at timestep $t$. \\
        $s_{\mathrm p}$ & Reverse step near which fragment additions are most likely. \\
        $\Delta_t$ & Number of nodes at the deletion boundary at timestep $t$. \\
        $f_\theta$ & Main denoiser producing $Z^X$, $Z^E$, and $Z^S$. \\
        $g_\phi$ & Auxiliary model predicting the boundary count $\Delta_t$. \\
        $\widehat P^S_{ij}$ & Predicted distribution over local attachment atoms in fragment $i$ for its edge to fragment $j$. \\
        $\mathcal P$ & Ranked pool of unique valid products returned by the assembler. \\
        $\widehat M$ & Final atom-level molecule selected from $\mathcal P$. \\
        \bottomrule
    \end{tabular}
\end{table}

\subsection{Fragment Graph Preprocessing}

Each molecule is decomposed into fragments using the preprocessing pipeline of
FragDiffusion.  In FragDiffusion, each connection between two fragments is represented by a single categorical code called an attachment mode. For an edge $(i,j)$, this code identifies the atom in fragment $F_i$, the atom in fragment $F_j$, and the bond type connecting them. For example, one attachment mode may represent ``connect atom 3 of $F_i$ to atom 1 of $F_j$ with a single bond.'' The possible attachment modes
and their empirical frequencies are obtained from the training data and stored
in a lookup table.

FragFlex uses this lookup table only during preprocessing to convert the
attachment-mode labels into neural attachment-site targets. For each
fragment edge $(i,j)$, preprocessing recovers the two local atom indices and
stores them as $S_{0,ij}$ and $S_{0,ji}$, while replacing the attachment-mode
edge with the binary labels
\[
E_{0,ij}=E_{0,ji}=1.
\]
Consequently, the model is trained and sampled using binary fragment topology
and explicit directed attachment-site targets. Unlike FragDiffusion, FragFlex
does not use an attachment-mode lookup table during inference; instead, its
neural attachment-site predictor directly proposes the atom-level connection
for each generated fragment edge.

\subsection{FragFlex Training}
\label{app:forward}

\begin{algorithm}[H]
\small
\caption{FragFlex training}
\label{alg:app_training}
\begin{algorithmic}[1]
\Require Dataset of clean fragment graphs $G_0=(X_0,E_0,S_0)$
\Repeat
  \State Sample a minibatch and $t\sim\operatorname{Unif}\{1,\ldots,T\}$
  \State Select a root $r\sim\operatorname{Unif}(V_0)$
  \State Sample $\tau_i\sim\operatorname{Cat}(\zeta')$ for each $i\neq r$;
         set $\tau_r=T+1$
  \State $V_t\gets\{i\in V_0:\tau_i\geq t\}$
  \State $B_t\gets\{i\in V_0\setminus\{r\}:\tau_i=t\}$
  \State Remove nodes with $\tau_i<t$ and their incident edges
  \State Mark nodes in $B_t$ and their incident edges as $\textsc{del}^{*}$
  \State Corrupt all other retained node and edge labels with $\bar Q_t$
         to obtain $G_t$
    \State Form the count-network input $G_t^-\gets G_t\setminus B_t$
    \State Set the count target $\Delta_t\gets|B_t|$
  \State $(Z^X,Z^E,Z^S)\gets f_\theta(G_t,t)$
  \State $Z^\Delta\gets g_\phi(G_t^-,t)$
  \State $\mathcal L_X\gets\operatorname{CE}(Z^X,X_0)$ over retained nodes
  \State $\mathcal L_E\gets\operatorname{CE}(Z^E,E_0)$ over retained node pairs
  \State Mask $Z^S_{ij,a}$ for $a\geq A_i$
  \State $\mathcal L_S\gets\operatorname{CE}(Z^S,S_0)$ over directed endpoints
         of clean fragment edges
  \State $\mathcal L_\Delta\gets
         \operatorname{CE}(Z^\Delta,\Delta_t)$ 
  \State $\mathcal L\gets\mathcal L_X+\lambda_E\mathcal L_E
         +\lambda_\Delta\mathcal L_\Delta+\lambda_S\mathcal L_S$
  \State Update $\theta$ and $\phi$ with AdamW
\Until{$N_{\mathrm{epoch}}$ epochs have been completed}
\end{algorithmic}
\end{algorithm}
\textbf{Forward corruption.} \quad
For each clean graph $G_0$, we sample a timestep
$t\sim\mathrm{Unif}\{1,\ldots,T\}$ and a root
$r\sim\mathrm{Unif}(V_0)$. Each non-root node receives a deletion time
$\tau_i\sim\mathrm{Cat}(\zeta')$, while $\tau_r=T+1$. Define the retained nodes $V_t$
and deletion-boundary nodes $B_t$ as
\begin{equation}
 V_t=\{i\in V_0:\tau_i\ge t\},
 \qquad
 B_t=\{i\in V_0\setminus\{r\}:\tau_i=t\}.
 \label{eq:app_retained_boundary}
\end{equation}
Nodes with $\tau_i<t$ are absent from $G_t$. Following GrIDDD, nodes in $B_t$ use the transient
state $\textsc{del}^{*}$, and every edge incident to a boundary node uses the
corresponding transient edge state. All other retained node and edge labels are corrupted with the empirical-marginal kernel,
\begin{equation}
 \bar Q_t(m)=\bar\alpha_t I+(1-\bar\alpha_t)\mathbf{1}m^\top,
 \label{eq:app_ordinary_kernel}
\end{equation}
where $m$ is the empirical fragment or edge marginal and $\bar\alpha_t$ is the
cumulative cosine schedule. The protected root uses the empirical fragment
marginal as its terminal prior. 
Thus, $G_T$ contains one fragment sampled from the empirical fragment
distribution and is not restricted to a single-atom fragment.

The resulting corrupted graph $G_t$ and the current timestep $t$ are passed to
the main denoiser $f_\theta$. The denoiser predicts logits for the clean
fragment identities, binary fragment-edge labels, and directed attachment
sites:
\[
(Z^X,Z^E,Z^S)=f_\theta(G_t,t).
\]
Here, $Z^X$, $Z^E$, and $Z^S$ denote the node, edge, and attachment-site logits,
respectively. For a source fragment $F_i$ containing $A_i$ atoms, positions
$a\geq A_i$ in $Z^S_{ij,a}$ are masked because they do not correspond to real
atoms.

\textbf{Fragment deletion and insertion schedule.} \quad
The structural schedule $\zeta'(t)$ determines when each non-root fragment is
removed during forward corruption or, equivalently, added during reverse
sampling. Because reverse sampling proceeds from $t=T$ to $t=1$, we define
$s_t=T-t$ as the number of reverse steps completed after the initial one-root
graph $G_T$. The parameter $s_{\mathrm{p}}$ specifies the reverse step near which fragment additions are most likely to occur.

We compute the unnormalized timestep weights as
\begin{equation}
u_t=\sigma(x_t)\bigl(1-\sigma(x_t)\bigr),
\qquad
\zeta'(t)=\frac{u_t}{\sum_{\tau=1}^{T}u_\tau},
\qquad
x_t=\frac{s_t-s_{\mathrm p}}{w\,T}.
\end{equation}
where $\sigma$ is the sigmoid function and $w\,T$ controls the temporal spread of
fragment additions. Each
non-root fragment then receives an event time
$\tau_i\sim\operatorname{Cat}(\zeta')$. We use $T=500$, $s_{\mathrm{p}}=0$, and $w=0.05$, concentrating graph growth near the beginning of reverse sampling.

\textbf{Fragment-count model.} \quad
At timestep $t$, the boundary set $B_t$ contains the fragments whose event time
is $\tau_i=t$. Its size $\Delta_t=|B_t|$
is the number of fragments that must be added at that step during reverse
sampling. The count network $g_\phi$ is trained to predict this number.

In the corrupted graph $G_t$, the boundary fragments are represented by
$\textsc{del}^{*}$ nodes. If these nodes were given directly to $g_\phi$, the
network could obtain the correct answer simply by counting them. We therefore
remove the boundary nodes from the count-network input:
\begin{equation}
G_t^-=G_t\setminus B_t.
\label{eq:app_count_target}
\end{equation}
The count network then predicts $\Delta_t$ from $G_t^-$ and the current
timestep:
\[
Z_t^\Delta = g_\phi(G_t^-,t),
\qquad
p_\phi^\Delta
=\operatorname{softmax}(Z_t^\Delta),
\]

where $p_\phi^\Delta$ is the predicted distribution over the possible values
of $\Delta_t$. Consequently, $g_\phi$ must infer how many fragments should be added from the
current graph structure and timestep, rather than reading the answer from
visible $\textsc{del}^{*}$ nodes.

\paragraph{Training objective.}
We use cross-entropy losses for fragment identity, edge, attachment-site, and
fragment-count prediction:
\begin{equation}
\begin{aligned}
\mathcal L_X
&=\operatorname{CE}(Z^X,X_0),
&\qquad
\mathcal L_E
&=\operatorname{CE}(Z^E,E_0),\\
\mathcal L_S
&=\operatorname{CE}(Z^S,S_0),
&\qquad
\mathcal L_\Delta
&=\operatorname{CE}(Z_t^\Delta,\Delta_t).
\end{aligned}
\label{eq:app_individual_losses}
\end{equation}
The first two losses are evaluated over retained nodes and node pairs,
respectively. The
total objective is
\begin{equation}
\mathcal L
=\mathcal L_X+\lambda_E\mathcal L_E
+\lambda_\Delta\mathcal L_\Delta+\lambda_S\mathcal L_S.
\label{eq_loss}
\end{equation}

\subsection{Flexible-Length Reverse Sampling}
\label{app:sampling}

\begin{algorithm}[H]
\small
\caption{FragFlex sampling}
\label{alg:app_sampling}
\begin{algorithmic}[1]
\Require Trained $f_\theta$ and $g_\phi$; empirical fragment marginal; $N_{\max}$
\State Sample one root fragment from the empirical fragment marginal to form $G_T$
\For{$t=T,T-1,\ldots,1$}
    \State Sample $\widehat\Delta_t$ from $g_\phi(G_t^-,t)$ and clip to the remaining capacity
    \State Append $\widehat\Delta_t$ nodes and mark them and their incident edges as $\textsc{del}^{*}$
  \State Predict clean node and edge logits with $f_\theta(G_t,t)$
  \If{$t=1$}
    \State Also compute and retain the attachment-site logits $Z^S$
  \EndIf
  \State Combine clean predictions with the exact categorical posterior $p_\theta(z_{t-1}\mid z_t,G_t, t)$
  \State Sample $X_{t-1}$ and the upper triangle of $E_{t-1}$; symmetrize $E_{t-1}$
  \State Connect each newly resolved node that has no neighbor in the pre-insertion graph
  \If{$t-1=0$ and the fragment graph is disconnected}
    \State Add highest-confidence cross-component edges until the graph is connected
  \EndIf
\EndFor
\State Mask padded site positions and convert the retained $Z^S$ to $\widehat P^S$
\State \Return $(\widehat X_0,\widehat E_0,\widehat P^S)$
\end{algorithmic}
\end{algorithm}

\textbf{Fragment insertion.} \quad
The count network $g_{\phi}$ predicts how many
fragment nodes should be added. Note that the
current graph has the same form as $G_t^-$, which is used to train the count
network. After sampling the predicted count $\widehat\Delta_t$, that many
$\textsc{del}^{*}$ nodes are appended, up to the maximum graph size
$N_{\max}$. These nodes are temporary placeholders. Their fragment identities
and edges are determined by the subsequent denoising step.

\textbf{Categorical denoising.} \quad
For each node or edge variable $z$, the main denoiser
$f_\theta(G_t,t)$ outputs clean-category logits defining
$p_\theta(z_0\mid G_t,t)$ after softmax.
This clean-category distribution is combined with the known
forward transition kernel to obtain the reverse distribution
\begin{equation}
p_\theta(z_{t-1}\mid z_t,G_t,t)
=
\sum_{z_0}
q(z_{t-1}\mid z_t,z_0)\,
p_\theta(z_0\mid G_t,t).
\label{eq:app_reverse_posterior}
\end{equation}

A new category $z_{t-1}$ is then sampled from this distribution.
Before sampling, probability assigned to the transient
$\textsc{del}^{*}$ state is removed and the remaining probabilities
are renormalized.

\textbf{Connectivity repair.} \quad
Because fragment edges are sampled independently, the generated fragment graph
is not guaranteed to be connected. If a newly inserted node is resolved into a
fragment without an edge to the existing graph, the sampler connects it to the
existing node with the highest predicted probability of a real edge. After the
final reverse step, any remaining components are joined by repeatedly adding
the highest-scoring edge between two different components. If the final graph
contains $C$ connected components before this repair, exactly $C-1$ edges are
added.

\textbf{Attachment-site prediction.} \quad
The attachment-site logits do not participate in the categorical reverse
posterior. They are therefore computed only at $t=1$, when the fragment
identities and topology are about to be resolved into $G_0$. Positions that do
not correspond to real atoms are masked, and the remaining logits are
normalized to produce the directed attachment-site distributions
$\widehat P^S$. These distributions are subsequently used by the neural
assembler to select atom-level connections.

\subsection{Valence-Aware Neural Assembly}
\label{app:assembly}

\begin{algorithm}[H]
\small
\caption{Neural top-$k$ valence-aware assembly}
\label{alg:app_assembly}
\begin{algorithmic}[1]
\Require Generated $(\widehat X_0,\widehat E_0,\widehat P^S)$;
         $k$, $B_{\mathrm{beam}}$, and $P_{\max}$
\State Extract the generated fragment-edge set $\mathcal E$
\ForAll{$(i,j)\in\mathcal E$}
  \State Construct and score the top-$k$ candidate pairs $\mathcal C_{ij}$
\EndFor
\State Order edges by increasing $|\mathcal C_{ij}|$
\State Initialize $\mathcal B$ with the disconnected RDKit fragments
\ForAll{edges $e$ in the selected order}
    \State Expand $\mathcal B$ over $\mathcal C_e$, retaining only
       RDKit-valence-safe branches
  \If{$\mathcal E$ contains a cycle}
    \State Add skip branches with penalty
  \EndIf
\State Rank by more successfully added bonds, fewer skips, a higher site
       score, and fewer non-top-1 choices
  \State Retain the best $B_{\mathrm{beam}}$ states in $\mathcal B$
  \If{$\mathcal B=\varnothing$}
    \State \Return failure
  \EndIf
\EndFor
\State Form $\mathcal P$ from the first $P_{\max}$ unique, connected,
       and sanitized products in beam order
\If{$\mathcal P=\varnothing$}
  \State \Return failure
\EndIf
\State Sample $\widehat M\sim\operatorname{Unif}(\mathcal P)$
\State \Return $\widehat M$
\end{algorithmic}
\end{algorithm}

Before top-$k$ selection, the assembler uses RDKit to exclude atoms that cannot
accept an external single bond, treating an explicit capping hydrogen as
replaceable. Here, $\widehat P^S$ denotes the attachment-site distribution
predicted by the neural network, and $\widetilde P^S$ denotes the distribution
obtained by applying this feasibility mask and renormalizing the remaining
probabilities.

For each generated fragment edge $(i,j)$, define
\begin{equation}
 \mathcal A_{i\rightarrow j}
 =\operatorname{TopK}_k(\widetilde P^S_{ij}),
 \qquad
 \mathcal A_{j\rightarrow i}
 =\operatorname{TopK}_k(\widetilde P^S_{ji}).
 \label{eq:app_site_candidates}
\end{equation}
The Cartesian product of these sets $\mathcal C_{ij}=\mathcal A_{i\rightarrow j}
  \times\mathcal A_{j\rightarrow i}$ contains at most $k^2$ candidate atom
pairs. For a candidate pair $(a_i,a_j)\in\mathcal C_{ij}$, the neural score is
\begin{equation}
\ell_{ij}(a_i,a_j)
=
\log\widetilde P^S_{ij,a_i}
+
\log\widetilde P^S_{ji,a_j}.
\label{eq:app_site_pair_score}
\end{equation}

Fragment edges are processed in increasing order of $|\mathcal C_{ij}|$, so
edges with fewer candidate atom pairs are considered first. For each edge,
every beam state is expanded over its candidate pairs. Each expansion replaces
an explicit cap hydrogen when needed, adds a single bond, and is retained only
if RDKit's strict valence check succeeds.

In a cyclic graph, an invalid predicted edge can be skipped because the
remaining edges may still keep the graph connected. Skipping is penalized and
is not allowed for trees, where every edge is required for connectivity. Beam
states are ranked by the following priorities: more successfully added fragment edges,
fewer skipped edges, a higher accumulated neural pair score, and fewer
non-top-1 attachment choices. After each fragment edge, only the best
$B_{\mathrm{beam}}$ states are retained. Thus, $k$ limits the local atom
choices for each edge, whereas $B_{\mathrm{beam}}$ limits the number of global
partial assemblies.

After all edges are processed, beam molecules are examined in the same ranking
order. They are fully sanitized, required to be connected, and deduplicated by
canonical SMILES. At most $P_{\max}$ unique valid products are
collected into a pool $\mathcal P$, and the final molecule is sampled as
\begin{equation}
\widehat M\sim\operatorname{Unif}(\mathcal P),
\qquad
|\mathcal P|\leq P_{\max}.
\label{eq_unif}
\end{equation}
Assembly fails if $\mathcal P=\varnothing$.

The search is score-based until $\mathcal P$ is constructed, while the final product is
sampled uniformly from this pool. This prevents the assembler from always
selecting the highest-scoring valid product and increases atom-level assembly
diversity for the same fragment graph. Because both the local candidate sets
and beam states are truncated, assembly may fail even when an unexamined valid
assembly exists.

\section{Additional Results}

\subsection{Scaffold Diversity and Molecular Quality}

We additionally evaluate scaffold-level diversity by counting unique
non-empty Bemis--Murcko (BM) scaffolds \citep{bemis1996properties}.
As shown in Table~\ref{tab:bm_scaffold_scaling}, models generating
more unique BM scaffolds tend to produce molecules with less favorable
QED and SA profiles, characterized by lower QED and/or higher SA
scores.
Although FragFlex does not achieve the highest scaffold diversity, it
obtains the best QED and SA scores while still generating millions of
unique BM scaffolds.
These results indicate that FragFlex provides a favorable balance
between scaffold diversity and molecular quality rather than
maximizing scaffold count alone.

\begin{table}[t]
\centering
\caption{
Molecular properties, distinct valid molecules, and unique
Bemis--Murcko (BM) scaffolds across generation scales.
QED and SA are evaluated at $n=10^5$.
Acyclic molecules are excluded from the BM scaffold counts.
}
\label{tab:bm_scaffold_scaling}

\begingroup
\footnotesize
\setlength{\tabcolsep}{2.5pt}
\renewcommand{\arraystretch}{1.08}

\begin{adjustbox}{max width=\linewidth,center}
\begin{tabular}{@{}lcc*{8}{r}@{}}
\toprule
& & &
\multicolumn{1}{c}{QED $\uparrow$} &
\multicolumn{1}{c}{SA $\downarrow$} &
\multicolumn{3}{c}{$|V\cap U|$ molecules $\uparrow$} &
\multicolumn{3}{c}{Unique BM scaffolds $\uparrow$} \\

\cmidrule(lr){4-5}
\cmidrule(lr){6-8}
\cmidrule(lr){9-11}

Model &
Unit &
Ins./del. &
\multicolumn{2}{c}{$n=10^5$} &
$10^5$ & $10^6$ & $10^7$ &
$10^5$ & $10^6$ & $10^7$ \\

\midrule

DiGress
    & Atom & No
    & 0.733 & 3.702
    & 71,458 & 712,891 & 7,113,016
    & 55,465 & 479,402 & 3,990,775 \\

GrIDDD
    & Atom & Yes
    & 0.651 & \underline{3.413}
    & 87,949 & 873,848 & 8,566,049
    & \textbf{77,285}
    & \textbf{691,875}
    & \textbf{5,889,181} \\

IFH
    & Atom & Yes
    & 0.725 & 5.367
    & 57,964 & 577,867 & 5,757,629
    & 57,093 & 558,945
    & \underline{5,402,044} \\

\midrule

JT-VAE
    & Fragment & No
    & \underline{0.762}
    &  3.418
    & \underline{98,975}
    & \underline{964,773}
    & 8,916,975
    & 66,996 & 504,974 & 3,404,902 \\

FragDiffusion
    & Fragment & No
    & 0.609 & 6.840
    & 96,073 & 952,672
    & \underline{9,265,827}
    & \underline{74,771}
    & \underline{618,016}
    & 4,734,767 \\

\textbf{FragFlex}
    & Fragment & Yes
    & \textbf{0.764}
    & \textbf{3.021}
    & \textbf{99,619}
    & \textbf{990,390}
    & \textbf{9,520,600}
    & 67,250 & 516,485 & 3,492,520 \\

\bottomrule
\end{tabular}
\end{adjustbox}

\endgroup
\end{table}

\subsection{Ablation Study}

We evaluate three design choices in FragFlex using $10^5$ generated
molecules.
The default FragFlex uniformly selects the final molecule from the
valid product pool $\mathcal P$ and uses a terminal graph $G_T$
containing a single fragment.
FragFlex-Top1 instead selects the highest-ranked product in
$\mathcal P$.
FragFlex-Middle shifts the peak insertion and deletion activity to the
midpoint, $s_{\mathrm p}=T/2$, following the schedule used by GrIDDD.
FragFlex-Multi allows the terminal graph $G_T$ to contain multiple
fragments, following the variable-size terminal-state formulation used
by GrIDDD.

The default FragFlex configuration achieves the highest number of
distinct valid molecules.
Selecting only the highest-scoring assembly candidate reduces
uniqueness, suggesting that uniform selection helps preserve molecular
diversity.
Moving the insertion and deletion peak to the midpoint slightly
improves validity but decreases uniqueness.
The multi-fragment terminal state achieves the highest uniqueness but
reduces validity.
Overall, the default configuration provides the strongest balance
between the two metrics.

\begin{table}[t]
\centering
\caption{
Ablation of the assembly selection rule, structural schedule, and
terminal graph at $n=10^5$.
$|V\cap U|$ denotes the number of distinct valid molecules.
}
\label{tab:fragflex_ablation}

\begingroup
\footnotesize
\setlength{\tabcolsep}{4pt}
\renewcommand{\arraystretch}{1.08}

\begin{adjustbox}{max width=\linewidth,center}
\begin{tabular}{@{}llrrr@{}}
\toprule
Variant &
Modification &
Validity (\%) $\uparrow$ &
Uniqueness (\%) $\uparrow$ &
$|V\cap U|$ $\uparrow$ \\

\midrule

FragFlex-Top1
    & $\widehat M=M^{(\mathrm{Top1})}$
    & 99.66
    & 99.22
    & 98,888 \\

FragFlex-Middle
    & $s_{\mathrm p}=T/2$
    & \textbf{99.76}
    & 99.32
    & 99,076 \\

FragFlex-Multi
    & Multi-fragment $G_T$
    & 99.40
    & \textbf{99.97}
    & \underline{99,370} \\

\textbf{FragFlex}
    & Default
    & \underline{99.67}
    & \underline{99.94}
    & \textbf{99,619} \\

\bottomrule
\end{tabular}
\end{adjustbox}

\endgroup
\end{table}

\section{Implementation Details}
\label{app:implementation}

\begin{table}[H]
    \centering
    \caption{Hyperparameters and model configuration}
    \label{tab:appendix_settings}
    \small
    \begin{tabular}{@{}ll@{}}
        \toprule
        Component & Setting \\
        \midrule
        Diffusion & $T=500$ with a cosine noise schedule $\bar\alpha_t$ ($s=0.008$) \\
        Deletion/insertion schedule
& Logistic $\zeta'(t)$ with $s_{\mathrm{p}}=0$, $w=0.05$\\
        Terminal prior $G_T$ & One fragment from the empirical fragment marginal \\
        Main denoiser $f_\theta$ & 7 layers, node width 384, edge width 96, 8 heads \\
        Count model $g_\phi$ & 2 layers, node width 128, edge width 32, 8 heads \\
        Maximum graph size & $N_{\max}=32$ fragments \\
        Training & 200 epochs, 4 GPUs, batch 32 per GPU, dropout 0.05 \\
        Optimizer & AdamW, LR $2\times10^{-4}$, weight decay $10^{-3}$ \\
        LR schedule & 5\% warmup from 0.1, cosine decay to 0.05 \\
        Assembly & $k=4$, $B_{\mathrm{beam}}=512$, $P_{\max}=16$, $\widehat M\sim\operatorname{Unif}(\mathcal P)$ \\
        \bottomrule
    \end{tabular}
\end{table}

Training was performed on four NVIDIA L40 GPUs. Training the
14.2M-parameter model for 200 epochs required approximately 10 hours of
wall-clock time. Sampling was performed on a single NVIDIA GeForce RTX 2080 Ti GPU, requiring approximately 3 hours to generate $10^{5}$ molecules.

\section{Code Availability}

Our code is available at
\url{https://github.com/SihyunPark99/FragFlex}

\clearpage
\section{Fragment Visualization}

\begin{figure}[H]
    \centering
    \includegraphics[
        width=1\linewidth,
        trim=0 0 0 0,
        clip
    ]{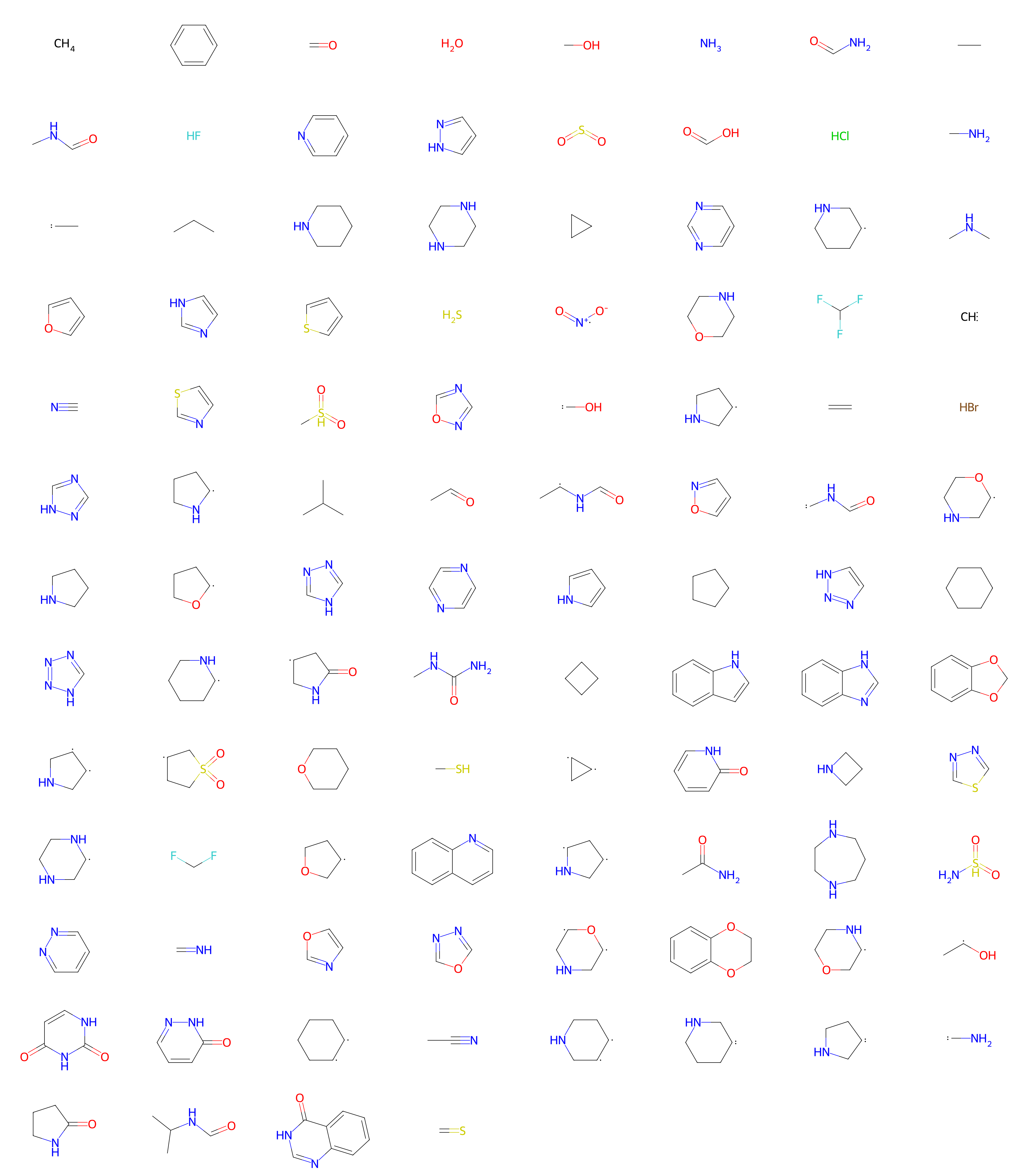}
    \caption{The 100 fragment types used by FragFlex, arranged in descending order of frequency in the ZINC training set.}
    \label{fig:fragment}
\end{figure}

\clearpage
\section{Generated Molecules}

\begin{figure}[H]
    \centering
    \includegraphics[
        width=1\linewidth,
        trim=0 0 0 0,
        clip
    ]{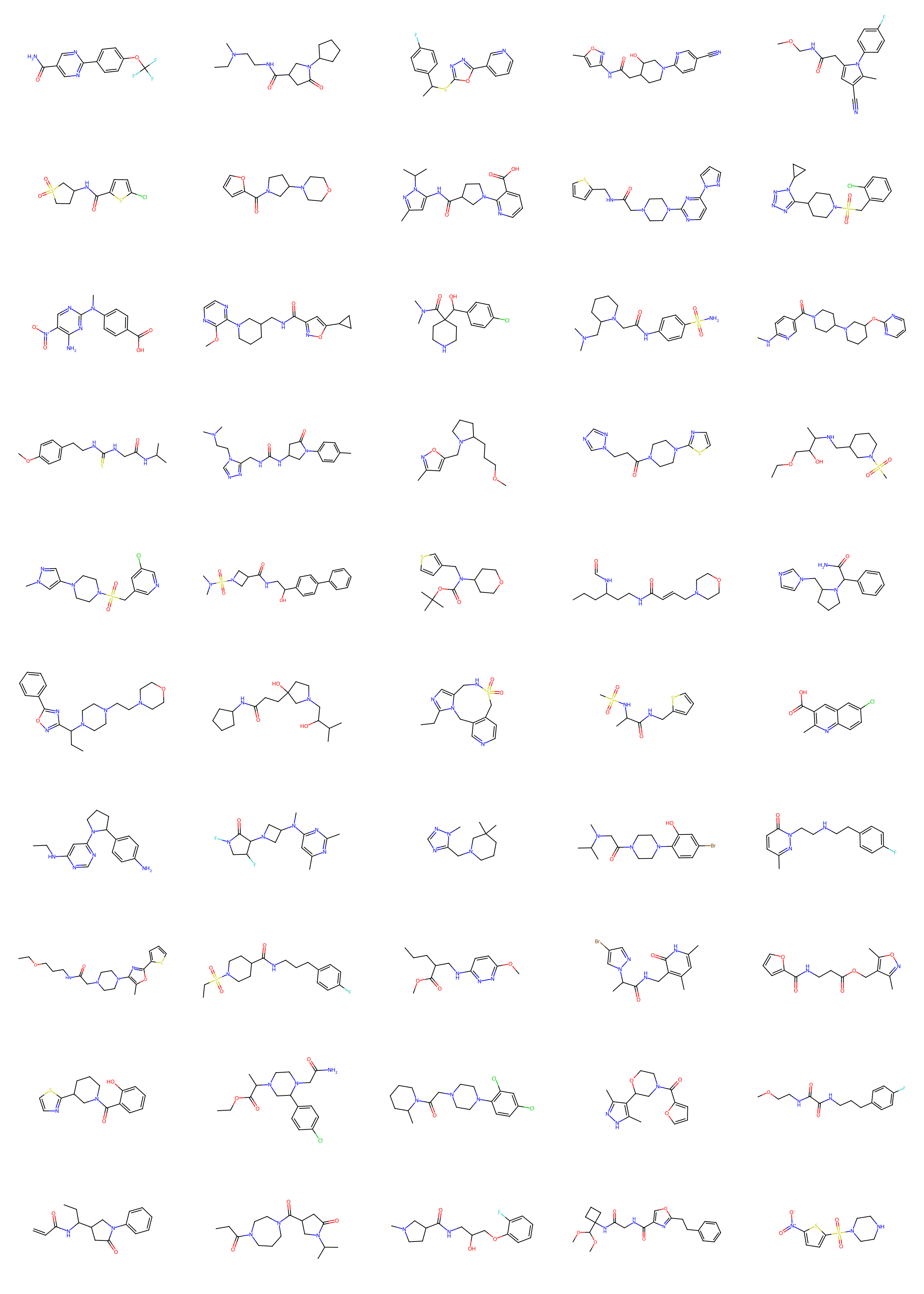}
    \caption{50 random samples generated by FragFlex.}
    \label{fig:generated-molecules}
\end{figure}

%%%%%%%%%%%%%%%%%%%%%%%%%%%%%%%%%%%%%%%%%%%%%%%%%%%%%%%%%%%%

\end{document}